%% file: main.tex
\pdfoutput=1

\documentclass[11pt]{article}

\usepackage[]{ACL2023}

\usepackage{times}
\usepackage{latexsym}

\usepackage[T1]{fontenc}

\usepackage[utf8]{inputenc}

\usepackage{microtype}

\usepackage{inconsolata}
\usepackage{tipa}
\usepackage{graphicx}

\usepackage{multirow}

\usepackage{booktabs}

\usepackage{amssymb}
\usepackage{pifont}

\usepackage{soul}
\usepackage{color}

\newcommand{\hlc}[2][yellow]{{\sethlcolor{#1}\hl{#2}}}
\definecolor{lavendergray}{rgb}{0.77, 0.76, 0.82}
\definecolor{lightgray}{rgb}{0.83, 0.83, 0.83}
\definecolor{whitesmoke}{rgb}{0.96, 0.96, 0.96}
\definecolor{timberwolf}{rgb}{0.86, 0.84, 0.82}
\newcommand{\cmark}{\ding{51}}%
\newcommand{\xmark}{\ding{55}}%

\def\comments{} 

\ifx\comments\undefined
    \newcommand{\lior}[1]
        {}
    \newcommand{\paul}[1]
        {}
    \newcommand{\roee}[1]
        {}
    \newcommand{\johan}[1]
        {}
    \newcommand{\orgad}[1]
        {}
    \newcommand{\idan}[1]
        {}
    \newcommand{\todo}[1]
        {}
\else
    \newcommand{\lior}[1]
    {{\color{blue}(Lior: #1)}}
    \newcommand{\paul}[1]
    {{\color{purple}(Paul: #1)}}
    \newcommand{\roee}[1]
    {{\color{orange}(Roee: #1)}}
    \newcommand{\johan}[1]
    {{\color{brown}(Johan: #1)}}
    \newcommand{\orgad}[1]
    {{\color{green}(Orgad: #1)}}
    \newcommand{\idan}[1]
    {{\color{pink}(Idan: #1)}}
    \newcommand{\todo}[1]
    {{\color{red}(ToDO: #1)}}
\fi

\usepackage{amsmath}
\usepackage{amssymb}
\usepackage{mathtools}
\usepackage{amsthm}
\usepackage[nameinlink]{cleveref}

\newcommand{\slpolicy}{\pi^{\mathrm{SL}}}
\newcommand{\rlpolicy}{\pi_{\theta}^{\mathrm{RL}}}
\newcommand{\anchorpolicy}{\pi^{\mathrm{anchor}}}

\newcommand{\vocab}{\mathbb{V}}
\theoremstyle{plain}

\theoremstyle{definition}

\theoremstyle{remark}

\DeclareMathOperator*{\NLI}{NLI}
\newcommand{\rNLI}{r^{\NLI}}
\newcommand{\rKL}{r^\mathrm{KL}}
\DeclareMathOperator{\kl}{KL}
\newcommand{\EOS}{{\scriptstyle \mathtt{[EOS]}}}

\newcommand{\RLEF}{RLEF}
\newcommand{\rllr}{\RLEF$_{\text{L}}$}
\newcommand{\rlhr}{\RLEF$_{\text{H}}$}
\newcommand{\GAE}{\mathrm{GAE}}

\DeclareMathOperator*{\argmax}{arg\,max}

\title{Factually Consistent Summarization \\ via Reinforcement Learning with Textual Entailment Feedback}

\newcommand{\authorspace}{\hspace{9pt}}

\author{Paul Roit$^{\beta\gamma}$$\thanks{~~Equal contribution}$ \authorspace Johan Ferret$^{\gamma*}$ \authorspace Lior Shani$^{\gamma*}$ \\
\AND
\bf Roee Aharoni$^{\gamma}$\authorspace Geoffrey Cideron$^{\gamma}$\authorspace Robert Dadashi$^{\gamma}$\authorspace Matthieu Geist$^{\gamma}$\\ 
\bf Sertan Girgin$^{\gamma}$\authorspace Léonard Hussenot$^{\gamma}$\authorspace Orgad Keller$^{\gamma}$\authorspace Nikola Momchev$^{\gamma}$\\
\bf Sabela Ramos$^{\gamma}$ \authorspace Piotr Stanczyk$^{\gamma}$\authorspace Nino Vieillard$^{\gamma}$\\
\AND
 Olivier Bachem$^{\gamma}$\authorspace Gal Elidan$^{\gamma}$\authorspace Avinatan Hassidim$^{\gamma}$\authorspace Olivier Pietquin$^{\gamma}$\authorspace Idan Szpektor$^{\gamma}$\\\\ {$^\beta$Bar-Ilan University}\authorspace{$^\gamma$Google Research}\\
 \footnotesize{\texttt{\{plroit,jferret,liorshani\}@google.com}}
 }

\begin{document}
\maketitle

\input{01_abstract}

\input{02_introduction}

\input{04_method}

\input{05_experimental_design}

\input{06_results}

\input{07_analysis_discussion}

\input{08_related_works}

\input{09_conclusions}

\input{96_limitations}

\input{97_ethics}
\input{98_acknowledgements}
\bibliography{custom}
\bibliographystyle{acl_natbib}

\appendix

\input{99_1_appendix.tex}
\input{99_2_appendix.tex}
\input{99_3_appendix.tex}

\end{document}

%% file: 01_abstract.tex
\begin{abstract}
    Despite the seeming success of contemporary grounded text generation systems, they often tend to generate factually inconsistent text with respect to their input.
    This phenomenon is emphasized in tasks like summarization, in which the generated summaries should be corroborated by their source article. 
    In this work we leverage recent progress on textual entailment models to directly address this problem for abstractive summarization systems.
    We use reinforcement learning with reference-free, textual-entailment rewards to optimize for factual consistency and explore the ensuing trade-offs, as improved consistency may come at the cost of less informative or more extractive summaries.
    Our results, according to both automatic metrics and human evaluation, show that our method considerably improves the faithfulness, salience and conciseness of the generated summaries.
\end{abstract}

%% file: 02_introduction.tex
\section{Introduction}

\begin{figure}[t!]
  \centering
  \includegraphics[width=\columnwidth]{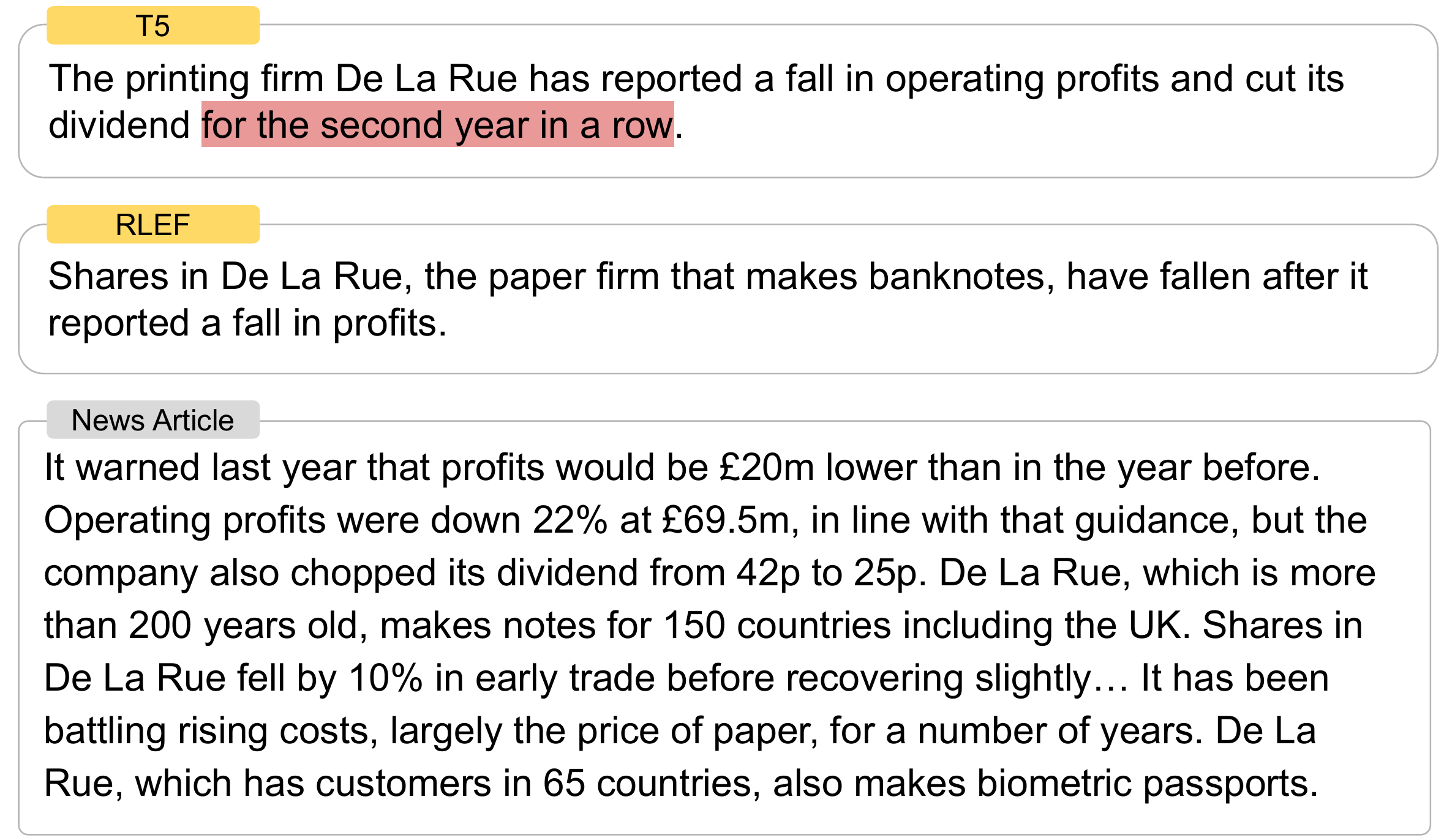}
  \caption{Summaries produced by multiple methods from a news article in the XSum dataset. Hallucinations or contradictions are highlighted in red. Note how the T5 generated summary mentions that there is a fall in operating profits \textit{for the second year in a row}, while the article only discusses a recent decline in earnings and a warning made in the previous year.}
  \label{fig:main_example}
  \vspace{-15px}
\end{figure}

Recent advancements in abstractive summarization systems~\citep{Zhang2019PEGASUSPW,liu-etal-2022-brio} are often impeded by their tendency to output information that is either contradicting or unsupported by their input article, often termed as ``hallucinations'' or factual inconsistency \cite{falke-etal-2019-ranking, maynez-etal-2020-faithfulness, pagnoni-etal-2021-understanding}. 
While these systems produce highly relevant and coherent text, this lack of factual consistency often limits their wide-spread adoption in real-world applications.
An example is depicted in \autoref{fig:main_example}, where the highlighted statement in the summary, while plausible, has no support in the input article.

Since widely-used metrics such as ROUGE \citep{lin-2004-rouge} were shown to be inefficient for detecting hallucinations, many recent research efforts introduced novel automatic metrics for measuring factual consistency \citep[inter alia]{kryscinski-etal-2020-evaluating, goyal-durrett-2020-evaluating, scialom-etal-2021-questeval}.
We propose to leverage these automatic metrics within a \emph{reinforcement learning} (RL) framework at training time.
Specifically, we apply \emph{textual entailment} assessment (a.k.a.\ \emph{natural language inference}, or NLI; \citealp{dagan-etal-2005,bowman2015large}) between the source article and the generated summary as a reward.

Our reward is based on the well studied textual entailment task \citep[][inter alia]{Pavlick2019InherentDI,McCoy2019RightFT,maccartney-manning-2007-natural}, for which there are many publicly available datasets~\cite{nie-etal-2020-adversarial,Liu2022WANLIWA}.
While these NLI datasets are not specific to summarization, it was shown that classifiers trained on these datasets perform well in detecting factual inconsistencies in summarization and other generative tasks \cite{honovich-etal-2022-true-evaluating}.
Because faithful summaries must be textually entailed from the corresponding input documents, using such a reward explicitly should guide a summarization model towards generating more factually consistent summaries. 
Yet, a high-quality summary should also be coherent and contain relevant information~\citep{fabbri-etal-2021-summeval}, aspects which may not be captured by entailment alone.
Moreover, a reward that is based only on entailment raises the risk of degenerate solutions, leading to either highly extractive~\citep{ladhak-etal-2022-faithful} or less informative summaries  \citep[``reward hacking'';][]{amodei2016concrete,skalse2022defining,pan2022the}.

To address these issues, we propose Reinforcement Learning with Entailment Feedback (\RLEF{}): Start with a model trained to produce summaries with the conventional cross-entropy objective, and further fine-tune it using RL with an entailment-based reward.
Throughout the RL procedure, we constrain the candidate models to stay close to the initial model. This way, while the model is being corrected for higher consistency, it also retains other summarization capabilities that were learnt with the maximum-likelihood (MLE) objective.
In this work we explore the consistent vs.\ informative trade-off in our RL-based summaries w.r.t.\ various aspects including model scale, regularization and decoding strategies. We find those aspects to be highly important and interdependent for the final model performance, highlighting the importance of carefully tuning them.

Our work stands in contrast to two prior RL-based approaches. 
The first approach induces a reward function from human feedback that encompasses various task-specific requirements into a single value~\citep{bohm2019better,Stiennon2020LearningTS}.
Collecting such feedback is expensive and requires dedicated data collection for each target task. 
In contrast, we use readily-available models and datasets for the reward, which address a specific aspect of generation that is generic across many different tasks.
Other works modeled the reward using different similarity functions between the \emph{reference} and the generated summaries~\citep{pasunuru-bansal-2018-multi,gunasekara2021using}, thus requiring reliable reference data.
Instead, our reward function evaluates the generated output only w.r.t.\ the \emph{input}, enabling to train using RL on data without reference summaries. 
We evaluated our approach on the widely used XSum~\cite{narayan-etal-2018-dont} dataset, using both automated metrics and human raters. The results show considerable improvements over strong baselines for factual consistency, salience, and conciseness of the generated summaries.

%% file: 04_method.tex
\section{Method}
\label{sec:method}

\begin{figure}
\includegraphics[width=\columnwidth]{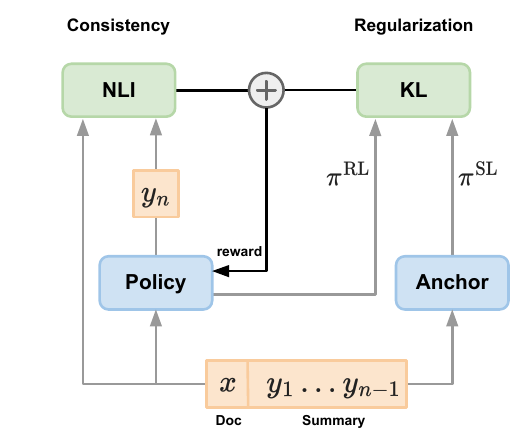}
\caption{RLEF training loop: (Left) given an input document, the policy generates a summary to be scored by the NLI model for consistency; (Right) given a document and the current generated summary, the KL distance between the RL and anchor model policies is used for regularization; Both scores are combined for training the policy. The black lines represents reward feedback to the model.
}
\label{fig:method}
\end{figure}
We would like to increase factual consistency using an entailment-based reward, while retaining the high salience and coherence that current summarization models already obtain.
To achieve this, we propose to initialize an RL policy with a summarization model trained on supervised data (the \emph{anchor model}).
From there, in each RL-based training step we update the parameters according to two signals: an entailment reward and a regularization term grounded on the anchor model.
During RL training, the entailment reward directs the model towards increased faithfulness, while the regularization term keeps the model from drifting to degenerate solutions and ``forgetting'' how to summarize.
 The process is illustrated in \Cref{fig:method}.

\subsection{\RLEF{}: RL from Entailment Feedback}
\label{sec:rl_in_detail}
\paragraph{Problem Formulation.}
We denote the input document and output summary as $x, y$ respectively. Let $\vocab$ denote the input and output vocabulary, and $y_{:n}=(y_1,..., y_n)$ denote the generated summary up to the $n$-th token.
We define the token-wise generative summarization process as a deterministic Contextual Markov Decision Process (CMDP, \citealt{hallak2015contextual}) with observable context, where the \textit{context} is the input text $x$, the \textit{state} at the $n$-th token generation is the sequence generated thus far $y_{:n-1}$, and the action space is defined over the vocabulary $\vocab$.
A policy $\pi(\cdot \mid y_{:{n-1}}, x) $, is a probability distribution over all tokens in $\vocab$, conditioned on the context and state.
We note that following this formulation, the policy is identical to a token-level auto-regressive language model~\cite{Bengio2003ANP}.
The RL objective is to find the optimal policy, which maximizes the cumulative reward signal. 

\paragraph{Rewards.}
We use an NLI classification model as a factual consistency reward signal. 
Since the model is trained to evaluate complete utterances and expects as input a \emph{grammatical} premise (document) / hypothesis (summary) pair, we use sequence-level rewards and define the token-level NLI reward to be zero on every token except for the end-of-sequence (EOS) token.
For the EOS token we set the reward to be the log-probability for an ``entailment'' decision according to the NLI classifier, using $x$ as the premise and $y_{:n}$ as the fully generated hypothesis:
\begin{equation*}
\rNLI(y_n ; y_{:n-1}, x)=
\begin{cases*}
	 \NLI(y_{:n};x) &  $y_n = \EOS$;\\
	 0 & otherwise,
\end{cases*}
\end{equation*}
where $\EOS$ is an end-of-sequence symbol, and $\NLI(y_{:n} ; x) = \log \Pr(\text{entailment}\mid y_{:n}, x)$.

To retain the summarization capabilities of the anchor model, we use Kullback-Leibler (KL) regularization to keep the RL-based policy close to the supervised anchor policy \cite{jaques2017sequence}:
\begin{align*}
   \rKL(y_n;y_{:n-1}, x) = \log \frac{\slpolicy( y_n \mid y_{:n-1},x) }{ \rlpolicy  (y_n \mid y_{:n-1}, x)}\ .
\end{align*}
 This term is added to the NLI reward, producing the final token-level reward:
\begin{multline}\label{eq: reward}
    r(y_n;y_{:n-1}, x) = 
           (1 -  \alpha) \rNLI(y_n;y_{:n-1},x)\\
           +  \alpha \rKL(y_n;y_{:n-1},x)\ .
\end{multline}
The hyperparameter $\alpha$ enables controlling the trade-off between enforcing faithfulness through the reward and remaining close to the anchor policy.

\paragraph{Training Algorithm.} We train the policy to optimize for the rewards defined in \Cref{eq: reward} using an on-policy actor-critic policy gradient (PG) approach. Since we keep proximity to the anchor model via the KL penalty reward, the algorithm can be considered a regularized PG algorithm, similarly to works by \citet{geist2019theory, shani2020adaptive,abdolmaleki2018maximum, tomar2020mirror, vaswani2021functional}; see \Cref{appendix:RL} for a detailed formulation.
Specifically, two models are learned: a policy (the generation model) and the expected value of the policy (the value network). We use the supervised model to initialize the parameters of both models, with the exception that the last layer of the value network outputs single scalars instead of a distribution over the vocabulary. 

The RL training process consists of the following stages: (1) Generating summaries with the current policy and (2) Scoring the summaries using the reward signal. Then, (3) Policy and value networks are trained,  jointly: the policy is trained via the PG loss while using the value for generalized advantage estimation (GAE, \citet{schulman2015high}); the value is trained via standard bootstrapping, using the GAE predictions.
Notably, this process does not require reference summaries for learning the policy. More details regarding the algorithm and losses can be found in \Cref{sec: Experimental Details Appendix}.

\subsection{Decoding at Inference Time}
As a direct consequence of RL training, the model explicitly learns to generate tokens with the goal of maximizing the long-term sequence reward. 
This is in contrast to MLE-based training, where the model learns to generate each token myopically, requiring heuristic decoding strategies such as beam-search to plan ahead.
As a result, we can use the more efficient temperature sampling instead of beam-search when decoding from an RL-trained policy.\footnote{We found that temperature sampling is sufficient for RL, while beam-search is required to improve the supervised policy.}

%% file: 05_experimental_design.tex
\section{Experimental Design}
\label{sec:experimental_design}
\subsection{Data}
We focus on XSum ~\citep{narayan-etal-2018-dont}, an abstractive summarization dataset that poses challenges around factual consistency. XSum is compiled from ~200K web-scraped BBC news articles, where the lead (introductory) sentence in every article is taken as the summary, and the rest of the sentences are taken as the source document.

Due to this formulation, XSum summaries may contain additional information that was not repeated in the rest of the sentences. 
Indeed, prior work found that only 20\% of the reference summaries in XSum are entailed from their source document~\citep{maynez-etal-2020-faithfulness}, and that summarization systems trained on XSum are likely to generate factually inconsistent summaries.
For this reason we find XSum suitable for our experiments, as we would like to see if the RL-based reward could alleviate the factual inconsistencies that supervised models learn to generate based on this data.

We also experiment on two additional datasets to compare to prior work. The TL;DR dataset~\citep{volske-etal-2017-tl}, using the same cleaned version provided by \citet{Stiennon2020LearningTS}, which contains ~120K Reddit posts and their short summaries, and the CNN/DM~\citep{nallapati-etal-2016-abstractive} dataset.
The latter contains ~200K news articles and their bullet-point highlights, which are mostly copied excerpts from article sentences. 
In this work we focus on abstractive summarization, and therefore evaluate our methods on CNN/DM with models trained, both supervised and reinforced, over TL;DR.

\subsection{Entailment Model}
In this work we focus on combining an existing entailment model as a reward in an RL framework.
We employ the NLI classifier from \citet{honovich-etal-2022-true-evaluating} across our study as a reward as well as for evaluation and data labelling for baseline methods.
It was trained over the ANLI dataset~\citep{nie-etal-2020-adversarial} with the T5-XXL architecture.
The classifier produces the characters `1' or `0' as its output for \emph{binary} entailment and non-entailment decisions, respectively.
We pose the source document as the premise and the predicted summary as the hypothesis, and use the log-probability of the decoded character `1' conditioned on the input as our reward.\footnote{We use an empirically validated classification threshold of 0.5 for entailment decisions.}
We leave improvements to the underlying factual consistency models for future efforts.
See \Cref{sec:related_work} for more discussion about different factual consistency models.

\subsection{Baseline Methods}

\paragraph{SL.}
Our supervised learning baseline is obtained by fine-tuning a T5 model on document-summary pairs.
We use the T5X framework~\citep{roberts2022t5x} for fine-tuning with batch size of 32 and keep the other hyperparameters to their default values (see \autoref{sec: Experimental Details Appendix} for details).
Fine-tuning is stopped once the model converges in terms of ROUGE on the validation set. 
This supervised baseline will also be used as the initialization checkpoint of our RL methods.
Decoding a summary using this model is implemented using beam search.

\paragraph{Filtered.}
Similar to the SL approach, with the distinction that we filter out training data where the summaries are not entailed by the input document according to our NLI model.
This filtering leaves ~60\% of the original XSum training set.
We train the model similarly to the SL model, and evaluate on the full validation and test splits, without filtering.
 
\paragraph{CTRL.}
Inspired by \citet{filippova-2020-controlled,rashkin-etal-2021-increasing}, we train the model on the full training set to explicitly differentiate between generating faithful and unfaithful summaries:
each training document is prepended with a phrase indicating if the target summary  is entailed or not according to our NLI model.
At inference, since we aim to produce consistent summaries, each document is always prepended with the phrase denoting an entailing summary, and continue decoding the summary using beam search.
Other parameters are similar to the SL method. 

\paragraph{FactPegasus.}
\citet{wan-bansal-2022-factpegasus} employ a tailored pre-training setup similar to PEGASUS~\cite{Zhang2019PEGASUSPW} that also takes factual consistency into account, and combine it with data pre-processing, and contrastive learning to generate more faithful summaries. 

\paragraph{CLIFF.}
\citet{cao-wang-2021-cliff} propose a contrastive learning objective that distinguishes between reference and heuristically created noisy summaries. 

\paragraph{RLHF.} \citet{Stiennon2020LearningTS} uses an RL approach with a reward model that learns from human comparisons of summaries. 
They iteratively add new feedback from humans for summaries generated by the current policy, and re-train the reward model.
We use their publicly released samples of the TL;DR validation set and the CNN/DM test set.

\subsection{Proposed Models}
We train two flavors of RL-based models. The first, \rllr{}, gives a lower weight to the regularization reward by setting $\alpha{}=0.1$ and the sampling temperature to 1. The second model, \rlhr{}, gives a higher weight to the regularization reward with $\alpha{}=0.2$ and a sampling temperature of 0.3. 
We altered both the $\alpha{}$ values and the sampling temperatures since we saw that both parameters affect the trade-off between factual consistency, as measured by the NLI model, and lexical similarity, as measured by ROUGE (see \Cref{fig:nli_vs_metrics}).
For additional implementation details see~\Cref{sec: Experimental Details Appendix}.

\subsection{Automatic Evaluation Metrics}
We report the common lexical n-gram overlap evaluation metrics and a set of factual consistency metrics, as the former were shown to be ill-suited for detecting unfaithful outputs~\citep{falke-etal-2019-ranking, pagnoni-etal-2021-understanding}.

For factual consistency, we report \textsc{NLI}, which is the percent of entailed summaries according to our NLI classifier, and the \textsc{$Q^2$} score \citep{honovich2021q}.
$Q^2$ is similar to QAGS \cite{wang-etal-2020-asking} and QuestEval \cite{scialom-etal-2021-questeval} but was shown to work better on XSum data \cite{honovich-etal-2022-true-evaluating} with higher correlation with human judgements.

When optimizing for faithfulness, an RL policy may resort to less abstractive summaries that are copied verbatim from the source~\citep{ladhak-etal-2022-faithful}, or less informative ones with a reduced level of detail.  
To explicitly measure these attributes in a summary, we report \emph{extractiveness} metrics: \textsc{Coverage} and \textsc{Density}~\citep{grusky-etal-2018-newsroom},
where the first measures the percent of summary tokens that also appear in the document, while the second measures a quantity similar to the average length of extractive spans in the summary.
Finally, we report the average summary length\footnote{We use $\texttt{SequenceMatcher::get\_matching\_blocks}$ from the python standard library to compute the set of extractive spans. Texts are tokenized with NLTK~\cite{loper-bird-2002-nltk}.} (\textsc{Length}).

\subsection{Manual Evaluation Protocol}
\label{sec:human_eval_protocol}
We asked human evaluators to rate a sample of the XSum test-set from several selected methods.
Each summary was evaluated by 3 different raters. Inspired by~\citet{fabbri-etal-2021-summeval}, we pose 4 questions outlining comprehensibility, attribution, salience and conciseness (see example in~\autoref{fig:raters_ui} in the appendix).
To get conclusive results, similarly to \citet{hannah2021measuring} we request binary yes/no answers and ask to answer ``No'' for any slight deviation from the desired property. 
For unfaithful summaries, the evaluator also provides the offending phrase. Our evaluator pool consists of 11 workers that successfully completed  a short training round of 10 examples (for details, see \Cref{sec:evaluator_details}).

%% file: 06_results.tex
\section{Results}
\label{sec:results}
\input{tables_figures/main_results}

\input{tables_figures/human_evaluation_table}

\paragraph{Automatic Evaluation.}
\Cref{tab:main_results} presents the automatic evaluation results on the XSum test set, comparing the supervised baselines to the two RL-based models (\rllr{}, \rlhr{}). 
 
The table shows that the RL-based models achieve the highest entailment scores as measured by the NLI and $Q^2$ metrics.
Notably, the RL approach is the most effective approach to utilize the NLI signal, scoring favorable compared to supervised baselines Filtered and CTRL, which leverage the same signal.

Analyzing ROUGE reveals the trade-off between the entailment and other summarization traits. Without strong regularization, \rllr{} scores highest on entailment but lower on ROUGE, indicating that in order to reach higher factual consistency, the model pushed farther away from the supervised starting point. The more strongly regularized \rlhr{} achieves a ROUGE score on par with the CTRL and SL baselines, suggesting that our KL-regularization prevented the policy from drifting.

Looking at extractiveness, the Density metric suggests that RL policies do not resort to copying text, and the increased Coverage implies that they tend to use more terms from the document, suggesting fewer hallucinations.
Lower ROUGE scores \emph{may} hint at lower quality summaries for the less regularized entailment model, yet the other metrics actually point at higher conciseness. 
We next present our human evaluation to shed light on these differences, and analyze whether the improvement in entailment is also captured by human readers, and whether the lexical divergence from the reference summary affects has implications on salience or conciseness.

\paragraph{Human Evaluation.}
The results of our human evaluation are detailed in \Cref{tab:human_eval_results}. Our raters fully agreed on 60\% of the examples regarding attribution. 
From attribution (factual consistency) perspective, the results strengthen the evidence that the RL approach is superior to other methods by a large gap. Interestingly the XSum reference summaries scored lowest with 23.6\%, showing that they are ill-suited to serve as faithful references for ROUGE and similar reference-based metrics. Notably, the human attribution evaluation was much stricter than the NLI metric, with much lower scores for all models, and we analyze this discrepancy in \Cref{sec:analysis}.

Surprisingly, the \RLEF{} models outperforms all other models also on Salience and Conciseness. Specifically, the less regularized \rllr{} learned to generate not only the most factually consistent summaries but also to improve on Salience and Conciseness, indicating that they are correlated w.r.t human quality perception.

\paragraph{Comparison with RLHF.}
We applied our RL approach on the TL;DR dataset.
We used the same input format and data split as in \citet{Stiennon2020LearningTS} for both the supervised and RL training processes.
For the supervised model (SL) we used hyper-parameters identical to our previous experiments (see \Cref{sec: Experimental Details Appendix}) except for a batch size of 128 and learning rate of 2e-4.

We compared our results using automated metrics with the RLHF approach \citep{Stiennon2020LearningTS}.
This approach is also based on the T5 model and uses a similar RL setup, yet it employs a reward model trained on \emph{task-specific human preferences} and applying a KL-based anchor. 
The results, detailed in \Cref{tab:tldr_results}, show that \RLEF{} achieves higher entailment scores in both NLI and $Q^2$ metrics, while our supervised model is on par with RLHF.
We also note that RLHF produces noticeably different and longer summaries compared to our supervised baseline, while \RLEF{} maintains similar length and ROUGE to the supervised baseline.

We also compared the two approaches in a transfer learning setting, where we predicted a summary on a different dataset (CNN/DM) using models trained on TL;DR. The results show similar trends, with higher entailment score for \RLEF{}.
These results hint at the benefit of utilizing a general NLI reward function, which managed to outperform the domain-specific RLHF reward both on the source domain and on a transfer setting.

%% file: tables_figures/main_results.tex
\begin{table}[t!]
\resizebox{\columnwidth}{!}{%
\begin{tabular}{llcccccccc}
\toprule
 & & \multicolumn{2}{c}{Faithfulness}              & \multicolumn{3}{c}{ROUGE}          & \multicolumn{3}{c}{Extractiveness} \\
Size & Method & NLI   & \multicolumn{1}{c}{$Q^2$} & 1   & 2 & \multicolumn{1}{c}{L} & Coverage    & Density   & Length   \\ \midrule
\multirow{5}{*}{XXL} & SL & 63.93 & \multicolumn{1}{l}{41.08} & \textbf{45.32} & 22.77 & \multicolumn{1}{l}{37.56}                 & 68.93       & 0.79      & \textbf{21.69}    \\
& Filtered & 74.54 & \multicolumn{1}{l}{43.01} & 43.84 & 21.36 & \multicolumn{1}{l}{36.24}                 & 69.21       & 0.81      & 20.74    \\
& CTRL  & 71.64 & \multicolumn{1}{l}{43.26} & 45.19 & 22.70 & \multicolumn{1}{l}{37.57}            & 69.83       & 0.82      & 20.94    \\
& \rllr{} & \textbf{94.66} & \multicolumn{1}{l}{\textbf{54.84}} & 41.77 & 19.95 & \multicolumn{1}{l}{34.75}                 & 75.03       & 0.98      & 17.72    \\
& \rlhr{} & 83.17 & \multicolumn{1}{l}{48.40} & 44.8  & 22.37 & \multicolumn{1}{l}{37.29}                 & 72.08       & 0.91      & 20.14    \\ \midrule
\multirow{4}{*}{Base} & SL & 52.44 & \multicolumn{1}{l}{36.16} & 39.84  & 17.77 & \multicolumn{1}{l}{32.63} & 71.77 & 0.87 & 20.52 \\ 
& \rllr{} & \textbf{79.90}  & \multicolumn{1}{l}{\textbf{46.70}} & 38.13 & 16.47 & \multicolumn{1}{l}{31.33} & 76.06 & 1.06 & 17.72 \\ 
& CLIFF & 68.16 & \multicolumn{1}{l}{45.71} & \textbf{45.17} & \textbf{23.32} & \multicolumn{1}{l}{\textbf{37.61}}                 & 73.37      & 1.21      & 20.86    \\
& FactPegasus & 62.01 & \multicolumn{1}{l}{42.69} & 37.16 & 15.13 & \multicolumn{1}{l}{30.36}                 &  \textbf{78.33}     & \textbf{1.42}      & 18.47     \\ \bottomrule
\end{tabular}%
}
\caption{Automatic evaluation results, XSum test set. \RLEF{} with various regularization patterns vs. baseline methods. Highest values are in bold. Due to stability issues in T5-Base RL-training (see \Cref{sec:analysis}), $T=0.3$ was used.}
\label{tab:main_results}
\end{table}

\begin{table}[t!]
\resizebox{\columnwidth}{!}{%
\begin{tabular}{clcccccccc} \toprule 
& & \multicolumn{2}{c}{Faithfulness} & \multicolumn{3}{c}{ROUGE} & \multicolumn{3}{c}{Extractiveness} \\
Test set & Method & NLI   & \multicolumn{1}{c}{$Q^2$} & 1   & 2 & \multicolumn{1}{c}{L} & Coverage    & Density   & Length   \\ \midrule
\multirow{3}{*}{TL;DR} & SL        & 94.11 & \multicolumn{1}{l}{74.34} & \textbf{36.75} & \textbf{14.87} & \multicolumn{1}{l}{\textbf{29.13}} & 91.40 & 3.86 & 27.69\\
& \rllr{}   & \textbf{99.39} & \multicolumn{1}{l}{\textbf{77.55}} & 36.58 & 14.81 & \multicolumn{1}{l}{29.12} & 92.89 & 4.14 & 26.57 \\
& RLHF-6B & 94.56 & \multicolumn{1}{l}{74.19} & 33.68 & 11.86 & \multicolumn{1}{l}{25.49} & 89.22 & 3.56 & \textbf{37.12} \\
\midrule 
\multirow{3}{*}{\begin{tabular}{c} CNN/DM \\ (transfer)\end{tabular}} & SL & 92.53 & \multicolumn{1}{l}{69.52} & 31.72 & 11.85 & \multicolumn{1}{l}{27.42} & 94.67 & \textbf{5.5} & 30.14 \\
& \rllr{} & \textbf{95.00} & \multicolumn{1}{l}{\textbf{71.08}} & 31.28 & 11.79 & \multicolumn{1}{l}{27.20} & \textbf{95.24} & 5.32 & 28.16 \\
& RLHF-6B  & 91.48 & \multicolumn{1}{l}{70.42} & \textbf{32.51} & \textbf{11.93} & \multicolumn{1}{l}{\textbf{27.85}} & 93.10 & 4.85 & \textbf{32.73} \\
\bottomrule
\end{tabular}%
}
\caption{Automatic evaluation results for TL;DR and CNN/DM test sets. Highest values are in bold.
\RLEF{} models are based on T5-XXL. 
For CNN/DM (transfer) we employ the \RLEF{} and SL models trained on TL;DR and predict summaries on the CNN/DM test-set, similarly to the \emph{transfer} setting in \citet{Stiennon2020LearningTS}.
For RLHF, we use the publicly available predictions of their human feedback model in the \emph{transfer} setting. }
\label{tab:tldr_results}
\end{table}

%% file: tables_figures/human_evaluation_table.tex
\begin{table}[t]
\centering
\resizebox{\columnwidth}{!}{%
\begin{tabular}{@{}llcccc@{}}
\toprule
 Size &Method & \textsc{Comprehension} & \textsc{Attribution} & \textsc{Salience} & \textsc{Conciseness} \\ \midrule
\multirow{4}{*}{XXL} & SL &  \textbf{99.0} $\pm$ 1.1   &   27.3 $\pm$ 5.0  &  61.6 $\pm$ 5.5  &  35.0 $\pm$ 5.4 \\
& Filtered &  96.3 $\pm$ 2.1   &   31.3 $\pm$ 5.2  &  61.3 $\pm$ 5.5  &  34.3 $\pm$ 5.3  \\
& \rllr{} &  98.7 $\pm$ 1.3   &   \textbf{56.6} $\pm$ 5.6  &  \textbf{78.0} $\pm$ 4.7  &  \textbf{61.0} $\pm$ 5.5  \\ 
& \rlhr{} &  98.0 $\pm$ 1.5   &   39.0 $\pm$ 5.5  &  70.6 $\pm$ 5.1  &  45.3 $\pm$ 5.6  \\ \midrule
\multirow{2}{*}{Base} & \rlhr{} &  96.0 $\pm$ 2.2   &   \textbf{38.3} $\pm$ 5.5  &  \textbf{64.3} $\pm$ 5.4  &  \textbf{44.3} $\pm$ 5.6  \\
& CLIFF  & \textbf{99.3} $\pm$ 0.9   &   28.3 $\pm$ 5.1  &  58.3 $\pm$ 5.6  &  33.3 $\pm$ 5.3  \\ \midrule
XSum & reference & 99.3 $\pm$ 0.9   &   23.6 $\pm$ 4.8  &  62.6 $\pm$ 5.4  &  30.3 $\pm$ 5.2  \\ \bottomrule
\end{tabular}%
}
\caption{Human evaluation results over 100 test set samples, each summary rated by 3 workers, results are micro-averaged. Each value corresponds to a percentage of positive answers per category with 95\% confidence intervals around the sample proportion. Highest values for each model size are in bold.
}
\label{tab:human_eval_results}
\end{table}

%% file: 07_analysis_discussion.tex
\section{Analysis}
\label{sec:analysis}

\begin{figure*}[t!]
    \centering
    \resizebox{\textwidth}{!}{
        \includegraphics[width=\textwidth]{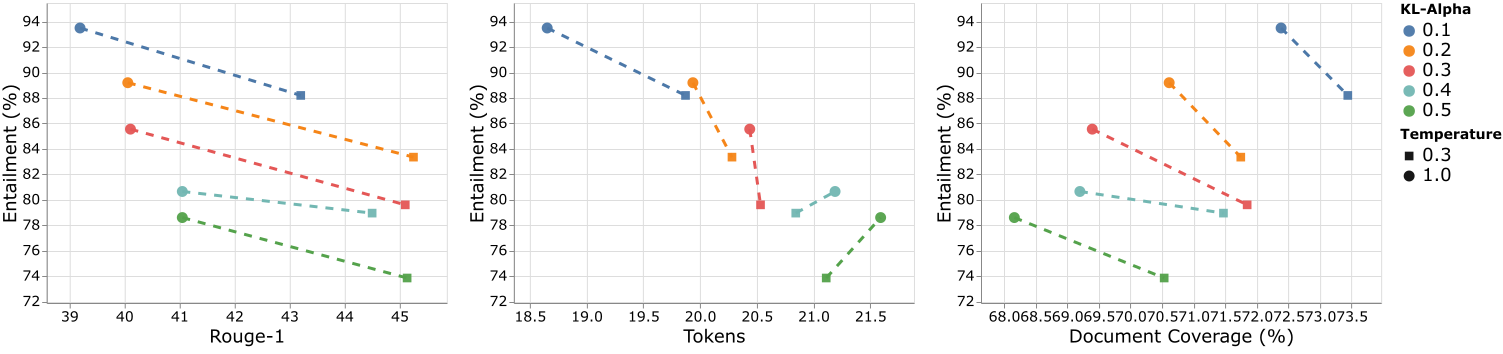}
    }
    \caption{Trade-offs between entailment and Rouge-1, Summary Token Length and Document Coverage as measured over the XSum validation set.
    Setups differ in KL-regularization (color) and sampling temperature (dot shape), model architecture is fixed to T5-XXL.
  }
    \label{fig:nli_vs_metrics}
\end{figure*}

\paragraph{Regularization and Sampling Temperature.} \Cref{fig:nli_vs_metrics} describes an ablation experiment where we vary the regularization $\alpha$ and the decoding temperature and measure the effect on different automatic metrics. 
Higher sampling temperature correlates with higher entailment and lower ROUGE scores.
We conjecture that this is since higher temperature generates more diverse summaries, which amplifies exploration away from the original gold references.
A similar phenomenon is observed when considering token length, as lower temperature policies produce summaries closer in length to the data-mean than their higher temperature counterparts.

As for the regularization coefficient $\alpha$, we observe the expected trade-off: lower regularization (smaller $\alpha$) leads to higher entailment (NLI), lower similarity to the supervised summary (ROUGE), and higher Coverage.
These may be explained by removal of external hallucinations that often use vocabulary terms that are unrelated to the document.

Surprisingly, in each KL setting, the lower temperature policy favors more document-aligned terms (perhaps for their higher initial probability), yet this is not reflected in the NLI metric, that stays lower than its higher-temperature counterpart.
We also observe that the summaries get shorter with less regularization, as the policy learns to mention fewer details as a way to alleviate generating inconsistencies.  

\begin{figure}[t!]
  \centering
  \resizebox{0.8\columnwidth}{!}{
  \includegraphics[width=\columnwidth]{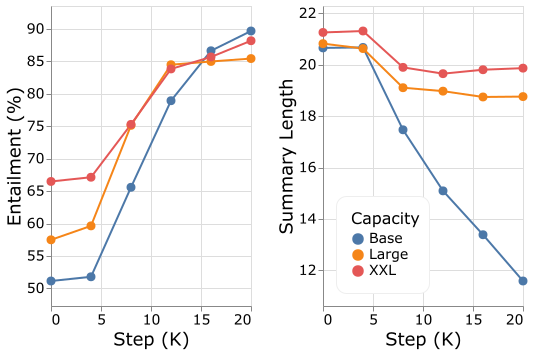}
  }
  \caption{Entailment ratio and summary length during RL training for different model sizes.}
  \label{fig:training_dynamics_across_t5_sizes}
\end{figure}

\paragraph{Model Size.} We tested our approach with different model sizes to study the effect of scale in the \rlhr{} setup.
We compared T5-Base (220M parameters), T5-Large (770M) and T5-XXL (11B), using the same hyper-parameters for all three models. \Cref{fig:training_dynamics_across_t5_sizes} shows the entailment rate on the XSum validation set during RL-finetuning. 
For all model sizes, our approach improved the entailment ratio over the supervised model by a large margin.

However, while the Large and XXL models changes the average summary length only slightly, the Base model completely degenerates, ``hacking'' the NLI reward by generating summaries that are half as short as the reference.
This suggests that higher-capacity models are essential to prevent reward hacking, perhaps due to two possible reasons.
First, the larger policies have higher generalization capabilities overall and can better accommodate different rewards, such as entailment and summarization regularization in our case.
Second, since the anchor model uses the same architecture, the higher capacity anchor model is more robust to changes in the summary and produces lower scores for less informative or more extractive summaries. 

\input{tables_figures/summaries_over_rl}

\subsection{Manual Analysis.}
To gain more insight into the inner workings of \RLEF{}, we propose two manual inspections about the types of changes being induced by the policy, and analysis of attribution errors found by our human evaluation procedure.

\paragraph{Changes to the summary during RL training.}
We study the changes that the \rlhr{} policy induces on a summary during RL training, focusing on the changes that cause a flip in entailment decision.
We sample 200 documents from the validation set for which we obtain the predicted summary at different checkpoints throughout the RL training process in 4K steps intervals.
We apply the NLI classifier for each document and summary list, and select 60 examples for which the NLI decision has flipped between any pair of consecutive checkpoints, and study what changes have been made to the summary that caused the flip.
Notably, most flips occur only once during training, and from the non-entailed to the entailed decision.
Examples are shown in \Cref{tab:summaries_over_rl_tuning} together with our categorization of the changes, with some summaries morphing in more  than one way.
We notice that for summaries produced by \rlhr{} most changes are local, meaning that the main predicate clause and the core participants remain the same throughout most checkpoints.
We classified 13 out of 60 examples as abstractively rephrased, where a specific detail is replaced with a broader description, e.g. \emph{returned to earth} instead of \emph{landed in Florida} (ex. 1).
However, we also found that 27 examples contained argument omissions, where verbal arguments or noun modifiers with typically non-core semantic roles~\citep{palmer-etal-2005-proposition} are removed (e.g. Locative or Temporal descriptions). See for example the ``Cause for arrest'' omission in ex. 5. 
Such omissions keep the information regarding the main participants intact, while lowering the risk of errors around non-core details.
Other changes included claim changes (16 cases) where a predicate has been replaced (see ex 3), argument replacements (8 cases), and other non-specific alterations.

\paragraph{Attribution error analysis.}
We analyzed attribution errors from the human evaluation of our best policy, \rllr{}, aggregated by majority vote. 
We inspect the offending phrase supplied by the evaluator for 39 out of 100 examples that are found to be non-attributable.
28 are considered as a local hallucination, mostly confirming to addition of personal names, numbers, places, and roles that did not appear in the article. 
For example, an article mentioned \emph{Kevin O'Malley} without alluding to his job title, while the summary referred to him as the \emph{Irish Ambassador}. 
While Kevin O'Malley was indeed an Irish ambassador, the model should not add such details if they are not explicitly mentioned in the article. 
Since most of these examples were found as entailing by our reward, this may point at issues with the NLI model that are due to knowledge conflicts between its parametric and contextual knowledge \cite{neeman2022disentqa}.
The rest of the examples include 5 contradictions and 5 major hallucinations.

%% file: tables_figures/summaries_over_rl.tex
\renewcommand{\arraystretch}{1.2} 
\begin{table*}[t!]
\centering
\scriptsize
\resizebox{\textwidth}{!}{%
\begin{tabular}{@{}p{0.005\textwidth}p{0.47\textwidth}p{0.47\textwidth}cp{0.09\textwidth}@{}}
\toprule
 & \footnotesize{Summary before the NLI flip} & \footnotesize{Summary after the NLI flip} & \footnotesize{NLI} & \footnotesize{Description} \\ \midrule
1  & Two astronauts who spent a year living on the International Space Station \hlc[lightgray]{have landed in Florida}. & Two astronauts who spent a year living on the International Space Station \hlc[lightgray]{have returned to Earth}. & {\color{OliveGreen}\cmark} & Abstractive Rephrasing  \\
2 & Afghan forces \hlc[lightgray]{have repelled an advance} by Taliban fighters on the northern city of Kunduz, officials say. & Afghan forces \hlc[lightgray]{have been battling} Taliban insurgents in the northern city of Kunduz.  & {\color{OliveGreen}\cmark} & Abstractive Rephrasing   \\
3 & A senior Nigerian military official has said militant Islamist group \hlc[lightgray]{Boko Haram is no longer a threat}, after a mosque attack that left at least 82 people dead. & A senior Nigerian official \hlc[lightgray]{has denied} that Islamist militant group \hlc[lightgray]{Boko Haram was behind a mosque attack} in the north in which more than 100 people were killed. & {\color{OliveGreen}\cmark} & Claim Change  \\
{4}  & Two people have been arrested \hlc[lightgray]{on suspicion of manslaughter} after a three-year-old boy died at a water park. & Two people have been arrested after a four-year-old boy died at a water park. & {\color{OliveGreen}\cmark} & Argument Omission \\
5  & Bolton Wanderers manager Lee Trotter has apologised after he and \hlc[lightgray]{striker Aaron Lennon} swore at fans on live television. & Bolton Wanderers manager Lee Trotter has apologised after he and \hlc[lightgray]{team-mate Gary Caldwell} swore at fans on live television. & {\color{red}\xmark} & Argument Change \\ \bottomrule
\end{tabular}%
}
\caption{Examples of summaries for the same document on consecutive checkpoints during RL training, before and after the NLI classification of the summary has flipped. The summaries maintain a stable main structure which enables manual inspection. The main changes are highlighted in gray, the NLI column depicts the entailment decision for the latter summary, and the description specifies the type of the semantic change.}
\label{tab:summaries_over_rl_tuning}
\end{table*}

%% file: 08_related_works.tex
\section{Related Work}
\label{sec:related_work}
\textbf{RL for text generation.} RL has been applied to many text generation tasks like neural machine translation~\citep{wu2018study, leblond2021machine}, extractive summarization~\citep{narayan2018ranking, wu2018learning, gao2019reward, arumae2019guiding}, abstractive summarization~\citep{chen2018fast} and others~\citep{bahdanau2016actor, welleck2019non, bai2022training, ouyang2022training, bai2022constitutional}.

Specifically for summarization, prior RL approaches used different reference-based metrics as a reward function.
In \citet{pasunuru-bansal-2018-multi}, two reward signals are measured between the generated and reference summaries: lexical overlap (ROUGE) to gauge salience and an entailment score to measure factual consistency.
\citet{gunasekara2021using} employed a similar approach with question-answering, they produced QA pairs conditioned on the generated summary to detect inconsistencies with the reference, and another set of QAs conditioned on the reference to measure salience.
Additionally, \citet{nan2021improving} proposed QUALS, a more computationally efficient QA approach, that was used in a contrastive learning setting. 
While their approach could be used without comparing outputs to reference summaries, they observed that adding such comparisons with the reference is essential for the stability of their method.
We note that for some datasets, reference summaries are likely to contain factual errors \cite{maynez-etal-2020-faithfulness}, decreasing the effectiveness of reference-based rewards.

Other RL methods, instead of explicitly defining the quality of a summary suggest to model it directly from human feedback~\citep{bohm2019better, ziegler2019fine, wu2021recursively, Stiennon2020LearningTS}.
This technique can prevent errors due to references that are misaligned with human judgment. 
While it is a promising approach, it also requires acquiring task-specific annotation, which can be labor-intensive.

Another hybrid approach interleaves a cross-entropy objective with policy gradients \cite{pang-etal-2021-agreesum} in multi-document summarization (MDS).
They use an in-domain NLI model, for which they annotate their MDS dataset with entailment decisions.
To stabilize their policy they employ an additional GAN-like training regime and add a discriminator loss between generated and reference summaries to their reward.

\textbf{Trade-offs in consistency models.} 
The choice of which factual consistency approach to use has interesting consequences for the RL setup.
Our work employs a binary NLI decision that does not point towards the specific inconsistent parts in the output summary.
Consequently, the reward is assigned to the final token of the summary, leaving proper credit assignment to the RL algorithm.
Other methods, specifically those based on question-answering \cite{durmus-etal-2020-feqa, wang-etal-2020-asking, honovich2021q} can frame misaligned answers in the generated summary and assign the reward explicitly to the offending tokens.
However, these QA-QG based methods may be much slower to compute.
Our reward requires a single forward pass using a transformer model over the document-summary pair, in comparison, QA-QG approaches require generating answer candidates, questions, answers from both sources and computing answer alignment.
Some of this complexity is remedied by generating jointly questions-and-answers \citep{nan2021improving}, but it still requires a lengthy decoding of QA pairs.
A different NLI-based approach decomposes the document and summary into smaller blocks of sentences \cite{laban-etal-2022-summac} and aggregates the final decision over a matrix of block-level NLI scores.
Such approach could aid the RL algorithm with credit assignment when generating long summaries.
In practice, the abstractive summarization datasets in this study use short single sentence summaries.

%% file: 09_conclusions.tex
\section{Conclusions and Future Work}
We propose to leverage NLI models as a ready-made, reference-free reward signal for RL training of factually consistent abstractive summarization models. 
Our experiments and thorough analysis with automatic and human evaluation show promising results for this approach, with our RL approach outperforming all baselines on factual consistency, while maintaining and even improving on other desired summarization attributes as well.

In future work, we would like to extend this approach to other grounded generation tasks, like knowledge-driven dialog.
In addition, we find it interesting to explore additional reference-free reward models for other summarization attributes (or for other tasks). Then, an important research direction would be to understand how to properly adapt our method to work with multiple such rewards.

%% file: 96_limitations.tex
\section*{Limitations}

While our approach shows promising results in both automatic and human evaluation, it relies on two significant pillars: a strong entailment model and a strong initial summarization model.
The NLI model implicitly encodes the biases and other data regularities that were part of the NLI training set into the generated summaries of our policy.
This is well demonstrated by the gap between human attribution judgements and the automatic NLI metric.
Our RL policies cannot improve on factual consistency errors if they are undetectable by the NLI reward. 
Hopefully, as NLI capabilities get better, so will the efficacy of \RLEF{} and the abilities to automatically flag hallucinations and contradictions.

Secondly, a strong summarization model is essential for our method in two ways: as an initialized starting point for RL exploration and as an anchor point to a policy.
While our RL training does not require any reference data and opens the possibility to use more un-summarized documents, it would probably not succeed as well without initializing from a high-quality supervised model.

Another limitation is that our experiments suggest that model size is important when using \RLEF{} (\Cref{fig:training_dynamics_across_t5_sizes}): both our summarization and NLI models are 11B parameters models. 
We believe it is important to further understand how to make our approach more robust to smaller models, to increase its computational efficiency and availability.

%% file: 97_ethics.tex
\section*{Ethics Statement}

Our work aims at solving the ethical issue of addressing misinformation in automated text generation tasks.
Yet, adopting automatic summarization by real users can amplify misinformation in cases where the model still makes an error or when the input text itself is not trustworthy.
As we stated in the limitations, our trained models heavily rely on other predictive models and therefore carry the biases of their training data, and may implicitly encode these into our generative process.
Therefore, we believe that to reach real-world use, not just our method should be scrutinized but also the NLI and summarization datasets that were used to train these models. 
Thus, such methods should be used with caution and combined with other techniques to ensure humans are capable of judging the validity of the information generated by the model.

%% file: 98_acknowledgements.tex
\section*{Acknowledgements}
We would like to thank our anonymous reviewers for their thorough comments and insightful suggestions.

%% file: 99_1_appendix.tex
\section{Experimental Details}\label{sec: Experimental Details Appendix}
\paragraph{RL Algorithm Details.}
We use an actor-critic on-policy PG algorithm with a learned value function $V_\psi$ and a parameterized policy $\pi_{\theta}$ to maximize the RL objective.
The policy gradient w.r.t. to the regularized reward $r(y_t ; y_{:t-1},x)$ defined in \Cref{eq: reward} is
\begin{align*}
&\nabla_\theta J(\theta)\\
&= \mathbb{E}_{x, y \sim \pi_\theta} \Big[ \sum_{t=0}^T \nabla_\theta \log \pi_\theta(y_t | y_{:t-1},x) G^{\alpha}_t \Big],
\end{align*}
where for brevity we denote $G^{\alpha}_t = \sum_{t'=t}^{T} r(y_t ; y_{:t-1},x)$, the accumulated regularized return. For more details on the derivation of this expression, and framing the regularized objective as an RL problem, we refer the reader to \Cref{appendix:RL}.

We use the value $V_\psi$ as a baseline, a state-dependent function that can be subtracted in the policy gradient without changing it. This leads to the following equivalent policy gradient
\begin{align*}
    &\nabla_\theta J(\theta) = \mathbb{E}_{x, y\sim \pi_\theta} \Big[ \sum_{t=0}^T \nabla_\theta \log \pi_\theta(y_t | y_{:t-1},x) \times \\
    & \quad \quad \quad \quad \quad\quad\quad\quad\quad\quad\quad\big(G^{\alpha}_t - V_\psi(y_{:t-1},x) \big) \Big]\\
    & = \mathbb{E}_{x, y\sim \pi_\theta} \Big[ \sum_{t=0}^T \nabla_\theta \log \pi_\theta(y_t | y_{:t-1},x) A^{\GAE}_\psi(y_{:t}) \Big]
\end{align*}
where $A_\psi$ is termed the advantage function. Applying this PG can be regarded a variant of the REINFORCE \cite{williams1992simple} algorithm with a baseline.
In practice, we replace the advantage in the expression above by \emph{generalized advantage estimation}~\citep[GAE,][]{schulman2015high}, which allows to better control the bias-variance trade-off via the $\lambda$ parameter:
\begin{align*}
    &A^{\GAE}_{\psi}(y_{t};y_{:t-1}, x) = \sum_{t'=t}^{+\infty} (\gamma \lambda)^{t' - t} \times \\  
    & \Big( r(y_{t'};y_{:t'-1}, x) + \gamma V_\psi(y_{:t'}, x) - V_\psi(y_{:t'-1}, x) \Big).
\end{align*}
Finally, the above policy gradient definition leads to the following per-example loss for learning the policy $\pi_{\theta}$,
\begin{align*}
    \mathcal{L}^{\pi}(\theta)(y_{:t},x) =  A^{\GAE}_\psi(y_{:t},x)\log \pi_\theta(y_t | y_{:t-1},x),
\end{align*}
where the gradients are only propagated here w.r.t. the policy parameters.

The value $V_\psi$ itself is learned via regression towards the return estimate induced by GAE, which is equivalent to minimizing the GAE advantage:
\begin{equation*}
    \mathcal{L}^V(\psi)(y_{:t},x) = \big(A^{\GAE}_{\psi}(y_{:t},x) \big)^2.
\end{equation*}
We now describe more intricate implementation details and hyper parameter choices.

\paragraph{RL Implementation Details.}

Given that we operate in the finite horizon setting, we naturally set the discount factor $\gamma$ to 1. 
Similarly to the PPO algorithm~\citep{schulman2017proximal}, we normalize the advantages in a given batch of data so that they approximately follow a standard normal distribution.
We also normalize the value loss by dividing it by the variance of the batch returns.
An important difference between our implementation and the standard (regularized) PG implementation is that instead of treating KL penalties along a given sequence as immediate rewards, we accumulate those and treat the resulting quantity as a sequence-level penalty. We found this to lead to more stability in the RL procedure. 

Unlike the conventional RL setting where both the policy and value are randomly initialized, in our case the policy is already fine-tuned to solve the required task. Thus, to make the value function accurate w.r.t.\ the already initialized policy, we observed that we needed a small number of iterations before the value estimation is sufficiently accurate to avoid detrimental policy gradients.
To do so, we run RL fine-tuning for 20K steps, with a warmup of 5K steps for the value network.
We also noticed that it was beneficial to use distinct values for the policy and value learning rates, so we decouple them in practice.

\paragraph{Optimization.}
We use Adafactor~\citep{shazeer2018adafactor} with a learning rate warmup phase: the learning rate is linearly annealed from zero to the specified asymptotic value.

\paragraph{Hyperparameter Search.}
We noticed that the optimal value of the policy and value learning rates are highly correlated.
Hence, we propose a decoupled hyperparameter search: we start by finding a suitable value learning rate by keeping the policy fixed. We then follow a standard grid search to find suitable values for the remaining hyperparameters including the policy learning rate, temperature and the regularization coefficient $\alpha$. Specifically, in our hyperparameter sweep we used temperatures $[0.1,0.3,1.0]$ and $\alpha$ values between $0.1$ and $0.8$ with a grid size of $0.1$.
Thus overall, our main sweep for the XXL model consisted of $24$ runs of $20K$ iterations.

We list all the hyperparameters used (unless different values are mentioned in the text) in Table~\ref{tab:rl-hparams}.
For the learning rate warmup and policy update delay, note that the number of steps reported correspond to gradient steps of the RL fine-tuning procedure.

\begin{table}
\centering
\begin{tabular}{lc}
\hline
\textbf{Hyperparameter} & \textbf{Value}\\
\hline
$\gamma$ & 1 \\
GAE $\lambda$ & 0.95 \\
Batch size & 32  \\ 
Temperature & 0.3 / 1.0  \\
Regularization $\alpha$ & 0.2 / 0.1 \\ 
LR warmup period & 2000  \\ 
Policy update delay & 5000  \\
Policy LR & 1e-5 \\
Value LR & 1e-5 \\
\hline
\end{tabular}
\caption{Hyperparameters for the RL fine-tuning procedure of \RLEF{}. When denoted left/right, left refers to the hyper parameter used for \rllr{} and right for \rlhr{}. }
\label{tab:rl-hparams}
\end{table}

\paragraph{SL Implementation details.}
For the SL models, we decode summaries with beam search with a beam width of 4 and a brevity penalty of 0.6.
For training we use the same optimizer with base learning rate of 0.001, batch size of 32, and a dropout rate of 0.1.

\paragraph{Resources.} We used TPU-v4 chips to train all the models mentioned. Each of our T5-XXL based \RLEF{} experiment ran for approximately 17 hours on 64 TPU chips. Furthermore, our main hyper parameters sweep included 24 such experiments, accounting for 1088 TPU-days.

%% file: 99_2_appendix.tex
\section{Evaluator Demographics, UI and Instructions}
\label{sec:evaluator_details}
\begin{figure*}[t!]
  \centering
  \includegraphics[width=\textwidth]{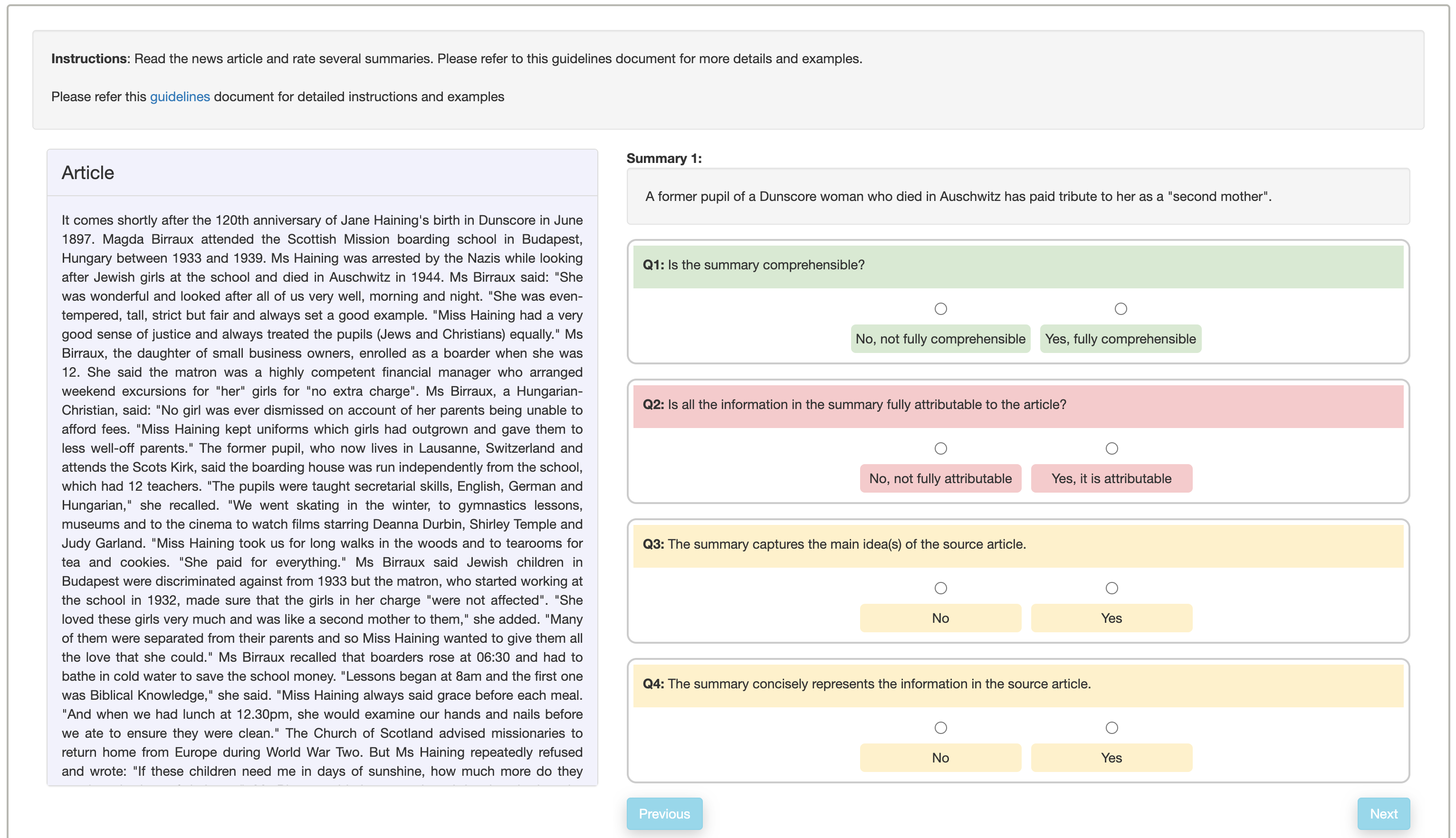}
  \caption{The evaluator user interface that we have used to rate summaries.}
  \label{fig:raters_ui}
\end{figure*}

We employed full-time hourly workers to rate the summary quality. 
Our raters consist of native English speakers, nationals from the U.S. and U.K. that hold graduate (70\%) and high-school (30\%) diplomas.
We supplied them with 2 pages of instructions and additional examples, and conducted an initial pilot study and training batch before proceeding to rate the summaries.
The UI that we used is displayed in \Cref{fig:raters_ui}.
In what follows we attach the guidelines presented to the raters in the human evaluation described in \Cref{sec:human_eval_protocol}. 
The guidelines are loosely based on \citet{hannah2021measuring}.

\subsection{Guidelines}

In this task you will be presented with a news article and multiple summaries of the article, and you are asked to evaluate the summary quality. You will rate each summary with 4 yes/no questions. These questions ask if the summary is:
Comprehensible and understandable.
Attributable (supported) by the article - no contradicting or unattested information.
Captures the main idea(s) behind the article.
Concise - does not contain additional details beyond the key information in the article.
Read carefully the text and the summary. The summaries may appear very fluent and well-formed, but contain slight inaccuracies that are not easy to discern at first glance.

\paragraph{Q1: Comprehensibility.}
An incomprehensible summary is not understandable due to significantly malformed phrases and sentences that are difficult to comprehend or make sense of. If there is any part of the summary that is unclear or hard to understand or malformed (e.g., partially cut-off or contains strange characters), select "No, not fully comprehensible".
Summary
When you leave it late, you leave it late is adding interest to your pension money as a result of the financial crisis. o, not fully comprehensible

\paragraph{Q2: Attributable (Supported) by the article.}
A fully supported summary contains information that can be found in the source article. No information in the summary is unattested when compared against the source news article. In other words, if you can say that “According to the news article…” with the summary following this phrase, you should answer, “Yes, it is attributable.” If some key details in the summary are not supported by the article (e.g. missing from the article), inaccurately represent the information in the article, or contradicted by the article, then please mark “No, not fully attributable.”

\paragraph{Q3: Main Idea.}
A main idea captures a fact or theme that is central to the article's discussion. It should involve the people, locations, or events that the article focuses on. If a main idea was removed from the original article, it would change the meaning, focus, or argument of the article. Note that this question is NOT asking whether the summary includes ONLY main ideas.

In Q3, to the best of your ability try to distinguish between the following cases, some may be more rare than others:
\begin{itemize}
    \item The summary is fully supported (yes to Q2) and captures the main idea (yes to Q3).
    \item The summary is fully supported (yes to Q2), but ignores the central point of the document (No in Q3).
    \item The summary contradicts the document in minor details or hallucinates some information (No to Q2), 
but the idea behind the document is mostly captured even if some details are incorrect (Yes to Q3).
\item The summary contradicts the document in key details (No to Q2) to the level where the main idea is unrecoverable or largely missed (No to Q3).
\end{itemize}

\paragraph{Q4: Conciseness.}
A summary is concise if it includes only the necessary details and the important information in the article. 
If it includes any details which are not central to the article, it should be marked as "No, it is not concise".
A summary may be concise even if some details are contradicting (i.e. you marked “No, it is not fully attributable” in Q2) as long as those were part of the main idea of the article.

In Q4 we are trying to find if the summary contains substantial information that does not belong to the main idea. 
If some minor details in the summary are contradicting, yet they are part of the main idea, then this summary is still concise, the system made an error of attribution, but not of over-generation.

%% file: 99_3_appendix.tex
\section{Fine-Tuning Language Models with Reinforcement Learning}\label{appendix:RL}

\subsection{Language Generation as a Contextual Markov Decision Problem}
\label{appendix:language-gen-mdp}
In this appendix, we explain the connection between arbitrary language generation tasks and the Markov Decision Process (MDP) framework~\citep{howard1960dynamic} which is widely used in RL.
We recall that an MDP $M$ is a tuple $M = (\mathcal{S}, \mathcal{A}, \gamma, r, P)$, where $\mathcal{S}$ is a state space, $\mathcal{A}$ is an action space, $\gamma \in [0, 1]$ is a discount factor, $r: \mathcal{S} \times \mathcal{A} \rightarrow [-r_{max}, r_{max}]$ is a bounded reward function and $P: \mathcal{S} \times \mathcal{A} \rightarrow \Delta_{\mathcal{S}}$ is a transition kernel. $\Delta_\chi$ denotes the standard simplex over $\chi$.
We represent sequential decision-making strategies as policies $\pi: \mathcal{S} \rightarrow \Delta_{\mathcal{A}}$. %
At any point in time $t$, a policy $\pi$ interacts in an MDP by observing the current state $s_t$, selecting an action $a_t \sim \pi(\cdot | s_t)$, and accordingly receiving a reward $r_t = r(s_t, a_t)$, before observing a new state $s_{t + 1} \sim P(\cdot | s_t, a_t)$.
We define the return as the discounted sum of rewards in one episode of interaction: $G_t = \sum_{t' = t}^{T} \gamma^{t'} r_{t'}$, where $T$ is called the horizon and is potentially infinite.
We now introduce Contextual MDPs (CMDPs)~\citep{hallak2015contextual}. They model the fact that a fixed context is available and determines the nature of rewards and dynamics. Formally, a CMDP is a tuple $M_c = (\mathcal{C}, f_M)$, where $\mathcal{C}$ is a context space and $f_M: c \in \mathcal{C} \rightarrow M$ is a function that maps a context to the corresponding MDP.

Any language generation task can be seen as the following interactive process: a language model observes the current state $s_t = y_{:t - 1}$ and context $c = x$, that is both the input text $x$ and the text generated so far $y_{:t - 1}$, and selects a token $a_t = y_t$. Thus, we can view any language generation task as a CMDP $M_c = (\mathcal{C}, f_M)$ with $f_M(c) = (\mathcal{S}, \mathcal{A}, \gamma, r(\cdot \, ; c), P(\cdot \, ; c))$, with the policy $\pi$ being the language model itself.
The state space $\mathcal{S}$ is the set of all potential generations (either complete or incomplete). We suppose that the maximum length of generated text $T$, which is equivalent to the horizon, and that of the input text $T_c$ are finite, which is a common assumption in NLP. Accordingly, if we note $\mathbb{V}$ the vocabulary (the set of all admissible tokens), we have $\mathcal{S} = \cup_{i=0}^{T} \mathbb{V}^i$.
Similarly, we have the context space $\mathcal{C} = \cup_{j=0}^{T_c} \mathbb{V}^j$.
The action space $\mathcal{A}$ is the set of tokens that the policy can output at any point in time, that is the vocabulary, hence $\mathcal{A} = \mathbb{V}$.
The discount factor $\gamma$ is arbitrary and can be set to 1 given that the horizon is supposedly finite.
The reward function $r$ is also arbitrary, but in the case of interest exposed in the main text we set it to:
\begin{equation*}
    r(s_t, a_t; c) = 
    \begin{cases}
        \NLI(y_{:t}; x) \text{ if } y_t = \EOS \text{ or } t = T, \\
        0 \text{ otherwise}.
    \end{cases}
\end{equation*}
Finally, and most importantly, the transition kernel is deterministic:
\begin{align*}
    &P(s_{t + 1} | s_t, a_t; c) = 
    \begin{cases}
        1 \text{ if } \EOS \in s_t \text{ and } s_{t + 1} = s_t, \\
        1 \text{ if } \EOS \notin s_t \text{ and } s_{t + 1} = y_{:t}, \\
        0 \text{ otherwise}.
    \end{cases}
\end{align*}
Indeed, any state that contains an $\EOS$ token can be considered an absorbing state.

\subsection{Language Generation From a Pre-Trained Model as a Regularized Markov Decision Problem}
\label{appendix:language-gen-reg-mdp}
While the previous formalism applies to all language generation tasks, we now describe a formalism that specifically applies to the language generation task that is explored in the main text: language generation when a pre-trained model is available.
It models the fact that we want generated text to be likely according to the pre-trained model, which we call \emph{anchor model} in what is next. We note the corresponding policy $\anchorpolicy$.
We consider the following reward function:
\begin{align*}
    r(s_t, a_t) &= (1 - \alpha) r(s_t, a_t) + \alpha \rKL(s_t, a_t),
\end{align*}
with the regularization term:
\begin{align*}
    \rKL(s_t, a_t) =  \log \anchorpolicy(a_t | s_t) - \log \pi(a_t | s_t) ,
\end{align*}
where $r$ is the reward function defined previously and $\alpha$ is a scalar controlling the regularization strength.
We recall that the Kullback-Leibler (KL) divergence between the current policy and the anchor policy has the expression:
\begin{align*}
    \kl(&\pi || \anchorpolicy)(s_t) = \\ &-\mathbb{E}_{a_t \sim \pi}\Big[\log \anchorpolicy(a_t | s_t) - \log \pi(a_t | s_t)\Big].
\end{align*}
Hence, the regularization term is an unbiased estimator for the KL divergence between current and anchor policies. 
Intuitively, it encourages the learned policy to keep a distribution that is close to the distribution over tokens induced by the anchor policy (the fine-tuned model).
Since the learned policy evolves along training, the reward function we described is non-stationary, that is the reward for a given state-action pair $(s, a)$ changes with the policy $\pi$.
Hence, the modified MDP is best viewed as a regularized MDP~\citep{geist2019theory}. 
We define the KL regularizer as $\Omega(\pi) = \kl(\pi|| \anchorpolicy)$, which is a strongly convex function.
We can show that this formalism is equivalent to the MDP with the non-stationary reward function described above.

\subsection{Defining the Reinforcement Learning Objective}
In this section, we show that the regularized reward defined in \Cref{eq: reward} can be used together with any PG based algorithm. To do that, we show that for any MDP (see~\Cref{appendix:language-gen-mdp}), the policy gradient can be easily re-derived for our regularization scheme when using parameterized policies. This repeats the derivations in \citet{schulman2017equivalence,geist2019theory}.
We denote trajectories $\tau = \{s_0\} \cup \{a_t, s_{t + 1}\}_{t=0}^{T-1}$. By a slight abuse of notations we denote the probability of a given trajectory under the policy $\pi$ as $\pi(\tau)$, that we can decompose as $\pi(\tau) = P(s_0) \prod_{i=0}^{T-1} \pi(a_i | s_i) P(s_{i+1} | s_i,a_i)$. We also denote $G_t$ as the return of a trajectory starting from time-step $t$.
Now, denote a parameterized policy $\pi_{\theta}$, and define the standard RL objective,
\begin{align*}
    J(\theta) &= \mathbb{E}_{\tau \sim \pi_\theta}\Big[\sum_{t=0}^T r(s_t, a_t)\Big], \\
              &= \mathbb{E}_{\tau \sim \pi_\theta}[G_0].
\end{align*}
The goal of RL is to find a parameterization $\theta^*$ that maximizes the following objective:
\begin{equation*}
    \theta^* \in \argmax_\theta J(\theta).
\end{equation*}
%
The policy gradient theorem states that
\begin{equation*}
    \nabla_\theta J(\theta) = \mathbb{E}_{\tau \sim \pi_\theta} \Big[ \sum_{t=0}^T \nabla_\theta \log \pi_\theta(a_t | s_t) G_t \Big].
\end{equation*}
We now place ourselves in the specific regularized MDP defined in \Cref{eq: reward} and \Cref{appendix:language-gen-reg-mdp}, with the reward regularization scheme, $r^{\alpha}(s,a) = (1 - \alpha) r(s, a) + \alpha \log{ \frac{\anchorpolicy(a | s)}{\pi_\theta(a | s)}}$. 
Define the RL objective of interest, which adds a regularization term to the reward function,
\begin{align*}
    J(\theta) =\mathbb{E}_{\tau \sim \pi_\theta} \Big[ \sum_{t=0}^T &(1 - \alpha) r(s_t, a_t) \\
    &+ \alpha \log \frac{ \anchorpolicy(a_t | s_t)}{\pi_\theta(a_t | s_t)}\Big].
\end{align*}

For $r(s_t, a_t)$, we repeat standard steps to re-derive the corresponding policy gradient. However, we need to have a separate treatment for the KL regularization reward $ \log\frac{\anchorpolicy (a_t | s_t)}{\pi_{\theta}(a_t | s_t)}$, as it explicitly depends on $\theta$.
We have:
\begin{align*}
    &\nabla_\theta \mathbb{E}_{\tau \sim \pi_\theta}\Big[\sum_{t=0}^T \log \frac{\anchorpolicy (a_t | s_t)}{\pi_{\theta}(a_t | s_t)}\Big] \\
    & = -  \nabla_\theta \mathbb{E}_{\tau \sim \pi_\theta}\Big[\sum_{t=0}^T  \log \frac{\pi_{\theta}(a_t | s_t)}{\anchorpolicy (a_t | s_t)}\Big] \\
    &=- \nabla_\theta \sum_\tau \pi_\theta(\tau) \sum_{t=0}^T \log \frac{\pi_{\theta}(a_t | s_t)}{\anchorpolicy (a_t | s_t)}, \\
    &=- \sum_\tau \nabla_\theta \Big( \pi_\theta(\tau) \sum_{t=0}^T \log \frac{\pi_{\theta}(a_t | s_t)}{\anchorpolicy (a_t | s_t)} \Big), \\
    &= -\underbrace{\sum_\tau \nabla_\theta \pi_\theta(\tau) \sum_{t=0}^T \log \frac{\pi_{\theta}(a_t | s_t)}{\anchorpolicy (a_t | s_t)}}_{A} \\ 
    &- \underbrace{\sum_\tau \pi_\theta(\tau) \nabla_\theta \sum_{t=0}^T \log \frac{\pi_{\theta}(a_t | s_t)}{\anchorpolicy (a_t | s_t)}}_{B}.
\end{align*}
We keep $A$ as is and show that $B$ is equal to 0:
\begin{align*}
    B &= \sum_\tau \pi_\theta(\tau) \nabla_\theta \sum_{t=0}^T \log \frac{\pi_{\theta}(a_t | s_t)}{\anchorpolicy (a_t | s_t)} \\
    &= \sum_\tau \pi_\theta(\tau) \sum_{t=0}^T \nabla_\theta \log \frac{\pi_{\theta}(a_t | s_t)}{\anchorpolicy (a_t | s_t)}, \\
    &= \sum_\tau \pi_\theta(\tau) \nabla_\theta \sum_{t=0}^T \log \pi_{\theta}(a_t | s_t), \\
    &= \sum_\tau \pi_\theta(\tau) \nabla_\theta \log \pi_\theta(\tau), \\
    &= \sum_\tau \nabla_\theta \pi_\theta(\tau), \\
    &= \nabla_\theta \sum_\tau \pi_\theta(\tau), \\
    &= 0.
\end{align*}
By putting all the pieces together we get the expression of the policy gradient for the modified RL objective:
\begin{align*}
    &\nabla_{\theta} J(\theta) \\
    \enspace &= \mathbb{E}_{\tau \sim \pi_\theta} \Big[ \sum_{t=0}^T \nabla_\theta \log \pi_\theta(a_t | s_t) \sum_{t'=t}^T r^{\alpha}(s_t,a_t) \Big],
\end{align*}
Denoting $G^{\alpha}_t$, the return of the trajectory when using $r^{\alpha}$, this can be rewritten as,
\begin{align*}
\nabla_\theta J(\theta) = \mathbb{E}_{\tau \sim \pi_\theta} \Big[ \sum_{t=0}^T \nabla_\theta \log \pi_\theta(a_t | s_t)  G^{\alpha}_t \Big].
\end{align*}
Note that we recovered the standard policy gradient for the regularized reward $r^\alpha$ (and corresponding return $G^{\alpha}$). This means that by treating $r^{\alpha}$ as the reward we can use any policy gradient method, to solve the new objective. Because this holds for any MDP, it holds for the specific MDP defined in \Cref{appendix:language-gen-mdp} for the summarization task. To see how this is concretely used in our approach to construct the PG losses, we refer the reader to \Cref{sec: Experimental Details Appendix}.